\documentclass{article}
\usepackage{arxiv}

\usepackage[utf8]{inputenc} % allow utf-8 input
\usepackage[T1]{fontenc}    % use 8-bit T1 fonts
\usepackage{hyperref}       % hyperlinks
\usepackage{url}            % simple URL typesetting
\usepackage{booktabs}       % professional-quality tables
\usepackage{amsfonts}       % blackboard math symbols
\usepackage{nicefrac}       % compact symbols for 1/2, etc.
\usepackage{microtype}      % microtypography
\usepackage{lipsum}		% Can be removed after putting your text content
\usepackage{graphicx}
\usepackage[numbers]{natbib}
\usepackage{doi}
\usepackage{amsmath}
\usepackage{subfig}
\usepackage{graphicx}
\usepackage{amsmath,amssymb,amsfonts}
\usepackage{algorithmic}
\usepackage{textcomp}
\usepackage{multirow}
\usepackage{makecell}
\usepackage{amsmath}
\usepackage{tabularx}
\usepackage{url}
\usepackage{xcolor}
\usepackage{mathrsfs}
\usepackage{amsmath}
\usepackage{setspace}

\DeclareMathOperator*{\argmax}{arg\,max}
\newcolumntype{P}[1]{>{\centering\arraybackslash}p{#1}}

\newcommand*\colourcheck[1]{%
  \expandafter\newcommand\csname #1check\endcsname{\textcolor{#1}{\ding{51}}}%
}

\newcommand*\colourcross[1]{%
  \expandafter\newcommand\csname #1cross\endcsname{\textcolor{#1}{\ding{55}}}%
}

\title{Boosting Few-Shot Learning with Disentangled Self-Supervised Learning and Meta-Learning for Medical Image Classification}

\author{\href{https://orcid.org/0000-0002-1321-9285}{\includegraphics[scale=0.06]{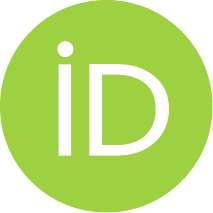}\hspace{1mm} Eva Pachetti\textsuperscript{1,2}}
\href{https://orcid.org/0000-0002-8795-9294}{\includegraphics[scale=0.06]{orcid.pdf}\hspace{1mm} Sotirios A. Tsaftaris \textsuperscript{3}}
\href{https://orcid.org/0000-0003-2022-0804}{\includegraphics[scale=0.06]{orcid.pdf}\hspace{1mm} Sara Colantonio \textsuperscript{1}}\\\\
\textsuperscript{1} Institute of Information Science and Technologies, National Research Council, Pisa, Italy \\
{\tt\small \{eva.pachetti, sara.colantonio\}@isti.cnr.it}\\
\textsuperscript{2} Department of Information Engineering, University of Pisa, Pisa, Italy \\
\textsuperscript{3} School of Engineering, University of Edinburgh, Edinburgh, UK\\
{\tt\small S.Tsaftaris@ed.ac.uk}}
\date{}

\renewcommand{\headeright}{}
\renewcommand{\undertitle}{Preprint submitted on 25 March 2024 to Elsevier}
\renewcommand{\shorttitle}{Disentagled Self-Supervision and Meta-Learning for Medical Images Few-Shot Classification}

\begin{document}
\maketitle

\begin{abstract}
\emph{Background and objective}: Employing deep learning models in critical domains such as medical imaging poses challenges associated with the limited availability of training data.
We present a strategy for improving the performance and generalization capabilities of models trained in low-data regimes. 

\noindent \emph{Methods}: The proposed method starts with a pre-training phase, where features learned in a self-supervised learning setting are disentangled to improve the robustness of the representations for downstream tasks. We then introduce a meta-fine-tuning step, leveraging related classes between meta-training and meta-testing phases but varying the granularity level.
This approach aims to enhance the model's generalization capabilities by exposing it to more challenging classification tasks during meta-training and evaluating it on easier tasks but holding greater clinical relevance during meta-testing. We demonstrate the effectiveness of the proposed approach through a series of experiments exploring several backbones, as well as diverse pre-training and fine-tuning schemes, on two distinct medical tasks,  i.e., classification of prostate cancer aggressiveness from MRI data and classification of breast cancer malignity from microscopic images. 

\noindent \emph{Results}: Our results indicate that the proposed approach consistently yields superior performance w.r.t. ablation experiments, maintaining competitiveness even when a distribution shift between training and evaluation data occurs. 

\noindent \emph{Conclusion}: Extensive experiments demonstrate the effectiveness and wide applicability of the proposed approach. We hope that this work will add another solution to the arsenal of addressing learning issues in data-scarce imaging domains.
\end{abstract}

% keywords can be removed
\keywords{Few-shot learning \and Meta-learning \and Disentangled representation learning \and Self-supervised learning \and Classification \and Medical image analysis }

\section{Introduction}
\label{sec:introduction}

\noindent Developing deep learning models that are able to perform competitively on small datasets is essential, especially in domains where data are scarce. Few-shot learning (FSL) is a paradigm designed to train models by using a limited number of examples \cite{Wang20}. One of the most relevant applications of FSL is in the medical image domain \cite{Carloni23,Dai23,Jiang22,Pachetti23}, where datasets are often small because of the difficulty and cost of acquiring such images and the need to protect patient privacy.
The use of FSL has become increasingly popular since the advent of meta-learning, defined as the ability to `learn to learn.` Indeed, in a meta-learning framework, models are trained on tasks rather than data, enabling the model to generalize better \cite{Chen19}.
Several studies \cite{Gidaris19,Su20} have shown that employing self-supervised learning (SSL) on additional unlabeled data as a pre-training step can boost the effectiveness of a meta-trained model. However, the main challenge with SSL methods, particularly contrastive-learning methods, is that they tend to disentangle features related to image augmentations used to generate additional instances for the contrastive setting (such as rotation and color jitter) rather than class-specific ones \cite{Wang21}. This way, models rely on the wrong features to perform the downstream task, resulting in poorer generalization. To learn more disentangled features beyond augmentations during SSL, Wang \emph{et al.} \cite{Wang21} proposed a new method called Iterative Partition-based Invariant Risk Minimization (IP-IRM). This iterative algorithm systematically disentangles one feature at a time by partitioning the training samples into two subsets based on an entangled feature (e.g., color) and minimizing a contrastive loss invariant w.r.t. the two subsets. The invariance between subsets ensures the disentanglement of the selected feature \cite{Wang21}. 
Our work builds on a simple idea: leveraging the power of disentangled SSL as a pre-training step to extract robust feature representations and coupling it with the generalization capabilities of a meta-learning framework to enhance the classification performance of models trained in the FSL regime.
Specifically, we use IP-IRM to pre-train a convolutional backbone to generate robust features and then fine-tune it with a meta-learning approach based on Prototypical Networks called Meta Deep Brownian Distance Covariance (Meta DeepBDC) \cite{Xie22}.
We also propose a novel meta-learning approach that uses different but related classes during the meta-training and meta-testing stages. Specifically, we use a dataset from the same domain with finer-grained classes for meta-training and coarser-grained classes for meta-testing, i.e., meta-training classes are subgroups of meta-testing classes. The aim is to train the model on more complex and diverse classification tasks, where complexity comes from higher granularity and diversity comes from a larger number of classes from which to choose in each episode. On the other hand, meta-testing episodes are easier to address since the classes are broader and the episodes less varied. This approach may be beneficial when coarse classes are still clinically relevant and there is insufficient data to expect good performance on more detailed classes.

To prove the versatility of our method, we evaluate our approach on two different medical imaging datasets and clinical tasks: the PI-CAI challenge dataset \cite{Saha23} of multiparametric MRI (mpMRI) prostate images and the  Breast cancer Histopathological Image (BreakHis) challenge dataset \cite{Spanhol15} of microscopic breast images. Using the first dataset, our objective is to classify the prostate mpMRI images based on the tumour severity by predicting a prognostic value defined by the International Society of Urological Pathology (ISUP) \cite{Egevad16}.  Following the mpMRI investigation, a biopsy examination (an invasive and uncomfortable procedure) is usually required to determine the tumour severity in suspected lesions \cite{Poppel21}. In this sense, a prediction model that performs accurate diagnosis directly from MRI images would help radiologists limit overdiagnosis, thus reducing patient discomfort. 

Concerning the BreakHis dataset, we consider as a clinical task the classification of benign and malignant lesions, which affects the patient's treatment plan. Breast cancer diagnosis from microscopic images is a challenging and time-consuming work as it requires manually detecting cancer nuclei \cite{George20}. This leads to high workload for pathologists, long diagnosis time and potential misdiagnosis due to human factors such as eye fatigue, in addition to device-dependant influences \cite{Benhammou20}. For all these reason, developing an automated classification of breast cancer biopsy samples could support pathologists by improving diagnosis accuracy and breast cancer early detection.

Below, we outline the primary contributions of our work:
\begin{itemize}
    \item We propose to exploit the strengths of disentangled SSL as a pre-training step for a meta-fine-tuned model to enhance the generalizability performance in the FSL domain.
    \item We propose a novel meta-learning approach that trains the model on more complex tasks than the ones used during meta-testing by employing related classes with different levels of granularity. Specifically, we use finer classes during meta-training and coarser ones during meta-testing, which are still clinically relevant. %more 
    \item We examine the robustness of our approach across distribution shifts by using data from different vendors between the training and evaluation phases.
\end{itemize}
We structured this paper as follows: Section \ref{sec:related_works} provides a comprehensive review of related works; Section \ref{sec:background} yields a theoretical background of the methods employed in this study; in Section \ref{sec:methods}, we delineate our proposed approach and detail the experimental settings, whereas the results of our experiments are presented in Section \ref{sec:results}. In Section \ref{sec:discussion}, we discuss and explain our findings. Eventually, in Section \ref{sec:conclusion}, we summarize our contribution and the applicability range of our study, finally outlining potential future directions.

%%%%%%%%%%%%%%%%%%%%%%%%%%%%%%%%%%%%%%%%%%%%%%%%%%%%%%%%%%%%

\section{Related works}
\label{sec:related_works}
\noindent Our work synergistically integrates meta-learning, SSL, and feature disentanglement methods to enhance the performance of models trained in an FSL setting. This section provides an overview of the state-of-the-art research in the interplay of these techniques, drawing parallels and highlighting differences with our proposed approach. First, we discuss existing works that leverage meta-learning methods to perform medical image classification in an FSL regime, aligning with our primary task. Next, similar to our approach, we examine studies that employ an SSL pre-training step to improve the performance of downstream tasks performed in an FSL setting. In the subsequent paragraph, we outline papers that propose different methodologies for improving FSL performance by leveraging feature disentanglement. Finally, we delve into the works that promote feature disentanglement in an SSL setting, culminating in a study that evaluates the efficacy of this approach in an FSL regime.
\subsection{FSL in Medical Image Classification}
\noindent Several studies have tackled the challenge of medical image classification within an FSL framework. Some of these works exclusively rely on meta-learning algorithms. For instance, Singh \emph{et al.} \cite{Singh21} applied the Reptile algorithm by meta-training the model on more prevalent diseases and evaluating it on rare diseases. Other works augmented the meta-learning paradigm by incorporating additional modules and pre-training steps. Dai \emph{et al.} \cite{Dai23}, for example, enhanced a gradient-based meta-learning algorithm by integrating it with a prior-guided Variational Autoencoder (VAE) to improve target features. From another perspective, Jiang \emph{et al.} \cite{Jiang22} merged meta-learning with transfer learning, utilizing a multi-learner model (autoencoder, metric-learner, and task learner) trained based on either a transfer learning or a meta-learning criterion at different stages. 
In this work, we propose to boost the meta-learning framework capabilities according to two main approaches: embedding a pre-training step that leverages disentangled SSL and elevating the generalization capabilities of the episodic meta-learning by varying the class granularity between meta-training and meta-testing episodes. 
\subsection{SSL as a pre-training step in FSL}
\noindent In line with our research, the concept of improving the performance of FSL-trained models through an SSL pre-training step has been explored in various studies. For instance, Chen \emph{et al.} \cite{Chen212} promoted a more robust data representation in downstream tasks by incorporating a contrastive SSL pre-training step followed by episodic training on natural images. In the facial expression recognition domain, Chen \emph{et al.} \cite{Chen23} integrated SSL and FSL by performing an SSL pre-training, followed by a classical fully-supervised fine-tuning phase of the feature extractor. The authors then used the fine-tuned model to build the prototype and perform a few-shot classification. In another approach Medina \emph{et al.} \cite{Medina20} adopted a transfer learning strategy that constructs a metric embedding, closely clustering unlabeled prototypical samples and their augmentations through a self-supervised approach. The pre-trained embedding serves as a starting point for few-shot classification, achieved by prototypical fine-tuning of the final classification layer. Yang \emph{et al.} \cite{Yang22} suggest incorporating contrastive learning into both the pre-training and meta-fine-tuning stages to enhance the performance of few-shot classification. In the pre-training stage, they introduced a self-supervised contrastive loss that leverages global and local information to learn effective initial representations. Concerning the meta-training stage, instead, they proposed a cross-view episodic training mechanism that involves performing nearest centroid classification on two different views of the same episode and employing a distance-scaled contrastive loss based on these views. Our work employs an enhanced SSL version for pre-training, which ensures the disentanglement of features, aiming at improving the quality of the features extracted during the subsequent downstream task.
\subsection{Feature disentanglement for FSL}
\noindent The concept of promoting feature disentanglement to improve performance in the FSL domain has been explored in the literature as well, particularly within the realm of natural images. Hu \emph{et al.} \cite{Hu23} employed feature disentanglement to generate augmented features through hallucination, aiming to mitigate data sparsity in few-shot classification. Similarly, the utilization of feature disentanglement for hallucination was adopted by Lin \emph{et al.} \cite{Lin21}, where they extracted class-specific and appearance-specific features for base categories and then employed them to hallucinate image features for novel categories. In another study, Cheng \emph{et al.} \cite{Cheng23} presented a disentangled feature representation framework tailored for few-shot applications. This framework adaptively decouples discriminative features, modeled by the classification branch, from the class-irrelevant component of the variation branch, thereby enhancing performance on few-shot tasks. 

Unlike the described approaches that employ feature disentanglement in a supervised setting to directly improve FSL classification, our work leverages the disentanglement of features at the unsupervised pre-training level. This unsupervised approach enhances the extraction of robust features, which subsequently benefits the FSL downstream task.
\subsection{SSL and Feature disentanglement}
\noindent Disentangled representation learning is a well-studied problem in supervised learning \cite{Zhu14,JT18,Cai19,Reed14,Karaletsos15,Liu22}. On the other hand, forcing the feature disentanglement in unsupervised learning is considered more challenging \cite{Locatello19}. Most of the existing methods for unsupervised disentanglement use generative models such as Generative Adversarial Networks (GAN) \cite{Chen16,Lin20,Ojha20} and VAE \cite{Chen18,Kim18,Higgins16}.
Wang \emph{et al.} \cite{Wang21} proposed the first method for unsupervised disentanglement called IP-IRM, which leverages the IRM algorithm \cite{Arjovsky19} to implement a disentangled SSL. Here, the authors assess the efficacy of their method in an FSL setting but differently from the present work, without fine-tuning the feature extractor on the downstream task. Indeed, they directly classify the feature embeddings generated by the feature extractor using both a k-nearest neighbors algorithm and a standalone linear classifier.

On the contrary, in this work, we consider the use of the IP-IRM algorithm as a pre-training step, intending to improve the learning of robust data representation for the downstream task, which is performed by fine-tuning the feature extractor through a meta-learning approach.

%%%%%%%%%%%%%%%%%%%%%%%%%%%%%%%%%%%%%%%%%%%%%%%%%%%%%%%%%%%%%%%%%%%%%%%%%%%%%%%

\section{Background and Motivation}
\label{sec:background}
\noindent Based on the state-of-the-art methods, we defined a new approach aimed at further improving the capabilities of models trained in an FSL manner. In this section, we offer a theoretical insight into the algorithms utilized for this purpose, motivating their relevance and adoption in our approach.

\subsection{SimCLR}
\noindent Zhang \emph{et al.} \cite{Zhang23} emphasized that for downstream classification tasks, predictive and contrastive SSL methods yield superior results compared to generative SSL approaches due to their ability to focus on high-level anatomical structures rather than pixel-wise information. Specifically, they demonstrated the robust performance of SimCLR, BYOL \cite{Grill20}, and RPL \cite{Doersch15} across various tasks, including segmentation and classification, highlighting their effectiveness in learning robust representations. Following these guidelines, we considered a contrastive method, namely SimCLR \cite{Chen20}, as our SSL algorithm. 

As demonstrated by \cite{Wang21}, a significant portion of the disentangled features learned by SimCLR are related to augmentations, whereas other meaningful features remain entangled. This results in a decreased generalization performance when evaluating the model on downstream tasks. To overcome this limitation, we employed the IP-IRM algorithm \cite{Wang21} to enforce the disentanglement of features beyond augmentation-related ones during SSL. In the subsequent paragraph, we provide a more detailed description of the IRM algorithm \cite{Arjovsky19} and its derivative IP-IRM algorithm \cite{Wang21}.

\subsection{Iterative Partition-based Invariant Risk Minimization}

\noindent It is often the case that during training the model environment-related features instead of class-specific ones. To identify and eliminate such features, we can leverage the assumption that if data are collected in different environments, spurious correlations are unstable, as they vary with changes in the environment itself \cite{Woodward05}.
Based on this principle, the IRM algorithm disentangles features by seeking a data representation where the optimal classifier remains consistent across all environments. In simpler terms, the same classifier performs effectively regardless of the environment of the training data, enabling robust generalization over out-of-distribution data. In summary, we aim for a data representation that is effective in producing accurate predictions and capable of generating an invariant predictor across all environments. Mathematically speaking, given a set of different environments $e$, a risk under the environment $R_e$, and a fixed and dummy classifier $w$ represented as a scalar, we want to find a representation of the data $\phi$ such that:
\begin{align}
    \min_\phi \sum_{e}R_e(\phi) + \lambda \cdot \lVert \nabla_{w|w=1} R_e(w\cdot\phi) \rVert ^ 2.
\end{align}

\noindent The first term corresponds to Empirical Risk Minimization (ERM) \cite{Vapnik92}, wherein the goal is to minimize the average risk across environments (predictive power) using a predictor $\phi$. The second term encourages the predictor $\phi$ to exhibit invariance across different environments. The parameter $\lambda$ represents a regularization term that balances these two components.

The authors in \cite{Wang20} leveraged this idea to force the disentanglement of features when the training is performed in an unsupervised, specifically in a self-supervised way, by iterating the IRM algorithm on different subsets of an unlabeled dataset.
Specifically, the idea is to subdivide the unlabeled dataset into two partitions based on a partition matrix $\mathcal{\textbf{P}}$. For the $i$-th image, $\mathcal{P}_{i,k} = 1$ if it belongs to the $k$-th subset and 0 otherwise. Each partition is then utilized to train the feature extractor $\phi$ in a self-supervised manner. The pretext task loss ($\mathcal{L}$) is defined by the contrastive loss, which takes the following form:
\begin{align}
\mathcal{L}= \sum_{x \in \mathcal{X}_k} - \log \left( \frac{\exp(x^T x^* \cdot  \theta)}{\sum_{x' \in \mathcal{X}_k \cup \mathcal{X}^* \backslash x} \exp(x^T x^* \cdot  \theta)} \right),
\end{align}

where $\theta$ = 1 is a parameter to evaluate the invariance of the SSL loss across the subsets, $\mathcal{X}_k$ represents the features of the $k$-th subset, and $\mathcal{X}^*$ the features of their augmented version.

\noindent In the first step, the parameters of $\phi$ are updated by:

\begin{align}
    \min_{\Phi} \sum_{\mathcal{\textbf{P}}\in \mathcal{P}} \sum_{k=1}^{2} 
     [ \mathcal{L} + \lambda_{1} \lVert \nabla_{\theta=1} \mathcal{L} \rVert ^ 2 ], 
\end{align}
where $\lambda_{1}$ is a hyperparameter that regulates the IRM loss term, with its minimization promoting invariance across subsets.

In the second step, the parameters of $\phi$ are held fixed, and the objective is to identify a new partition to be disentangled, denoted by $\mathcal{\textbf{P}^*}$. The desired partition is the one that maximizes the invariance between the losses, i.e.:

\begin{align}
    \mathcal{\textbf{P}^*} = \argmax_{\mathcal{\textbf{P}}} \sum_{k=1}^{2} \left [\mathcal{L} + \lambda_{2} \lVert \nabla_{\theta=1} \mathcal{L} \rVert ^ 2 \right].
\end{align}
The $\mathcal{P}$ matrix is updated as $\mathcal{P} \leftarrow \mathcal{P} \cup \mathcal{\textbf{P}^*}$. These two steps are iterated until convergence is achieved. For further details on the IP-IRM algorithm, refer to \cite{Wang21}.

\subsection{Meta Deep Brownian Distance Covariance}

\noindent Meta DeepBDC is a meta-learning method based on Prototypical Networks \cite{Snell17}. In traditional Prototypical Networks, data are represented by their first moment (mean), and similarity between class prototypes and query embeddings is assessed using metrics like Euclidean or cosine distance. However, studies have shown that incorporating richer statistics, including second moments, while adopting the Frobenius norm or Kullback-Leiberler divergence as similarity measures results in improved performance \cite{Wertheimer19, Li20}. It should be noted that these studies often neglect joint distributions, limiting the learning of relationships between the two embeddings. To overcome this limitation, Xie \emph{et al.} \cite{Xie22} proposed using the BDC metric, defined as the Euclidean distance between the joint characteristic function and the product of the marginals of two random variables, X and Y. BDC is mathematically formulated as:
\begin{align}
    \rho(X,Y) = \int_{\mathbb{R}^p}^{}\int_{\mathbb{R}^q}^{}\frac{|\Phi_{XY}(t,s)-\Phi_X(t)\Phi_Y(s)|^2}{c_pc_q||t||^{1+p}||s||^{1+q}}dtds,
\end{align}

\noindent where $\Phi_{X}(t)$ and $\Phi_{Y}(s)$ are the marginal distributions of $X$ and $Y$, respectively, $\Phi_{XY}(t,s)$ is the joint characteristic function of the two random variables and $c_p$ is defined as $c_p = \pi^{(1+p)/2}/\Gamma((1+p)/2)$, where $\Gamma$ is the complete gamma function.

In the image domain, the similarity between two images can be computed as the inner product between their respective BDC matrices. In the context of Meta DeepBDC, the algorithm calculates a BDC matrix for each support and query embedding. Subsequently, the prototype for each class is determined as the average of the BDC matrices belonging to that class. Classification is then executed by computing a distance distribution between the query BDC matrix and each class prototype. See \cite{Xie22} for more details.
 %%%%%%%%%%%%%%%%%%%%%%%%%%%%%%%%%%%%%%%%%%%%%%%%%%%%%%%%%%%%%

  \section{Material and Methods}
\label{sec:methods}

\noindent This section delves into the proposed approach, outlining the core concept and the experiments undertaken to validate its effectiveness.

\subsection{The main idea}
\noindent Our work aims to improve the capabilities of FSL-trained models by synergistically combining the strengths of SSL, feature disentanglement, and meta-learning methods. Specifically, we propose to perform a pre-training step exploiting an SSL setting where feature disentanglement is promoted, addressing the limitations of classical contrastive SSL. This step aims to equip the model with robust features that enhance its performance during the downstream task. To effectively disentangle features, we employ the IP-IRM algorithm. This algorithm iteratively partitions an unlabeled dataset into two partitions based on a feature to be disentangled. The disentanglement is performed by minimizing the invariance of the SSL loss across the two partitions.

\begin{figure*}
     \includegraphics[width=\textwidth]{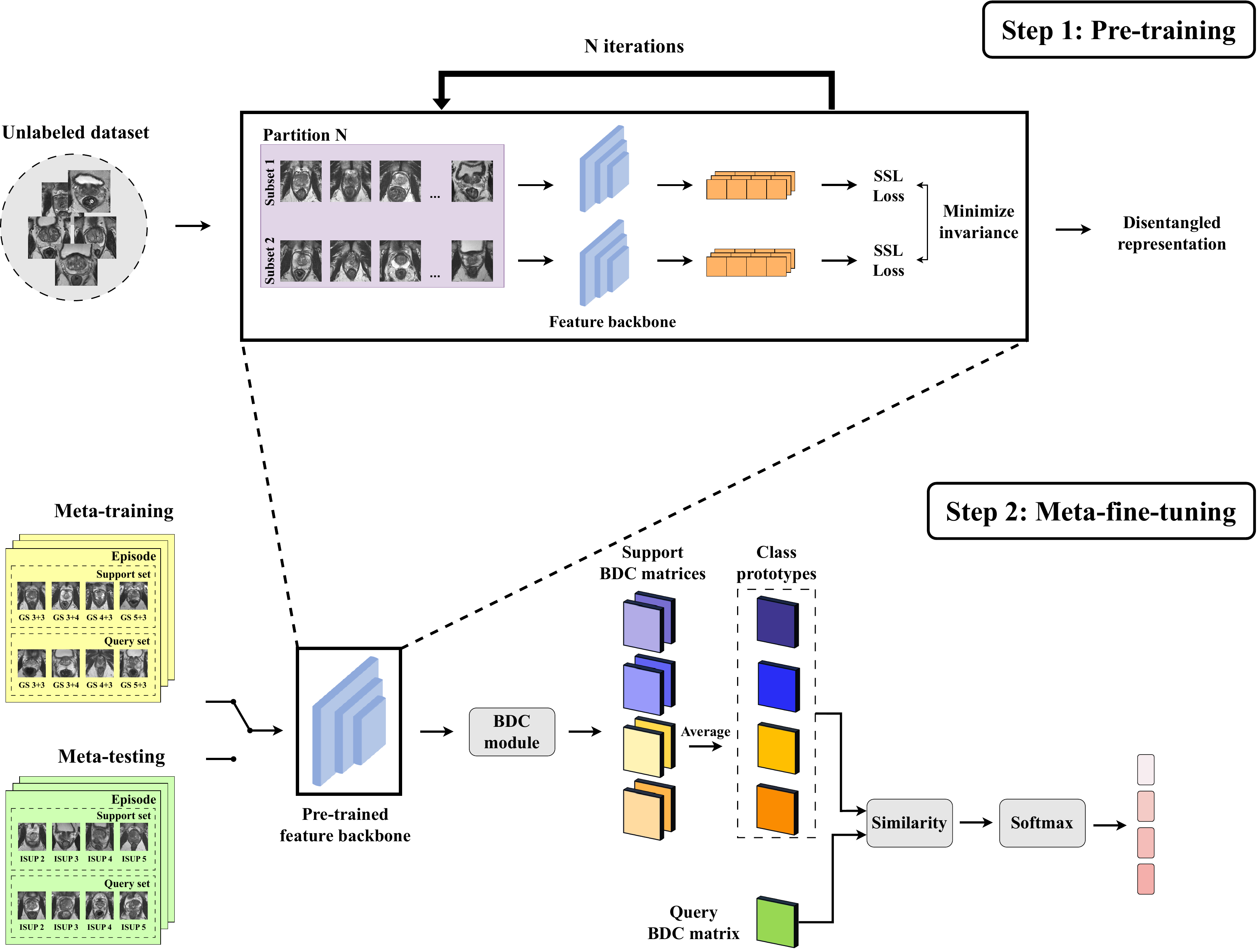}
         \caption{Illustration of the proposed approach. In the pre-training step, the feature backbone undergoes pre-training using the IP-IRM algorithm. At each iteration, the unlabeled dataset is divided into two subsets to maximize invariance between SSL losses. Subsequently, these subsets are employed to update the feature backbone parameters by minimizing the invariance between the SSL losses. The pre-trained backbone then undergoes meta-fine-tuning using the Meta DeepBDC algorithm. Meta-training episodes contain finer-grained classes while meta-testing coarser-grained ones belonging to the same source dataset. In both the meta-training and meta-testing phases a BDC matrix is computed for each support and query sample. Class prototypes are derived by averaging the BDC matrices of all support samples for that class. Classification is achieved by computing a similarity distribution of the query BDC matrix w.r.t. the class prototypes.}
         \label{fig:proposed_approach}
\end{figure*}

Subsequently, we employ meta-learning to fine-tune the pre-trained model on the downstream task using a labeled dataset. Meta-learning further enhances the model's generalization abilities, enabling effective performance in low-data settings. Specifically, we divide the labeled dataset into three subsets to perform meta-training, meta-validation, and meta-test. To further improve the meta-learning framework's effectiveness, we propose utilizing related classes at different granularity levels between the meta-training and meta-testing episodes, i.e., the meta-training subset is composed of images labeled ad a finer-grained level, whereas meta-validation and meta-test images are labeled with coarser-grained ones. 
We illustrate our proposed approach in Fig. \ref{fig:proposed_approach}.

\subsection{Problem definition}
\noindent Given a set of datasets \(D = \{D^1, D^2, \dots, D^N\}\), each one comprises both labeled and unlabeled images, denoted as \(D^i = \{(x, y)_1, \dots, (x, y)_m,(x)_{m+1}, \dots, (x)_n\}\), where \((x, y)_k\) represents an image-label pair, and \((x)_h\) is the image alone. We split each dataset as  \(D^i_{lab} = \{(x, y)_1, (x, y)_2, \dots, (x, y)_m\}\) and 
\(D^i_{unlab} = \{(x)_{m+1}, (x)_{m+2}, \dots, (x)_n\}\).
We further separate the labeled dataset into two distinct sets of classes, namely \(C^i_{fine}\) and \(C^i_{coarse}\), where \(|C^i_{coarse}| < |C^i_{fine}|\). Specifically, \(C^i_{fine}\) contains classes that are subgroups of classes in \(C^i_{coarse}\). In line with this, we further divide \(D^i_{lab}\) into three subsets for training, validation, and testing, labeling the images in each subset as follows: \(D^i_{meta-train} = \{(x, y)_k, y_k \in C^i_{fine}\}\), \(D^i_{meta-val} = \{(x, y)_k, y_k \in C^i_{coarse}\}\), and \(D^i_{meta-test} = \{(x, y)_k, y_k \in C^i_{coarse}\}\).

Our objective is to utilize \(D^i_{unlab}\) to pre-train a feature extractor \(\phi\) following an SSL approach, where feature disentanglement is enforced. Subsequently, we conduct fine-tuning on \(\phi\) using a meta-learning approach, leveraging \(D^i_{meta-train}\), \(D^i_{meta-val}\), and \(D^i_{meta-test}\).

%%%%%%%%%%%%%%%%%%%%%%%%%%%%%%%%%%%%%%%%%%%%%%%%%%%%%%%%%%%%%%%%%%%%%%%%%%%%5
\subsection{Pre-training phase}
\label{sec:methods_pre_training}

\subsubsection{Proposed pre-training step}
\noindent The initial step in our approach involves a disentangled SSL pre-training phase, which leverages the SimCLR and IP-IRM algorithms. As with the IP-IRM algorithm, our pre-training phase is conducted iteratively. Specifically, at each iteration, the $D^i_{unlab}$ is subdivided into two subsets w.r.t. a feature to be yet disentangled, and the SSL loss invariance is minimized between the two subsets.

\subsubsection{Pre-training ablations}

\noindent To evaluate the efficacy of the initial stage of our proposed approach, we compared it with two alternative pre-training methodologies. The first approach involves the use of the conventional SimCLR algorithm, corresponding to the initial iteration of the IP-IRM algorithm, where all samples belong to the first subset (i.e., the first column of $\mathcal{\textbf{P}}$ is set to 1). In this scenario, pre-training was conducted on the same dataset used by the IP-IRM algorithm, specifically $D^i_{unlab}$. The second approach entails fully-supervised pre-training on the Imagenet dataset, utilizing pre-trained parameters provided by the PyTorch library \cite{Paszke19}.

\subsection{Fine-tuning phase}
 
\subsubsection{Proposed Meta-learning framework}
\noindent After pre-training the feature extractor $\phi$, we meta-fine-tuned it on the downstream task using the Meta DeepBDC algorithm \cite{Xie22}, which employs an episodic training approach. During the episodic training, the model is presented with multiple classification tasks, each comprising a support set and a query set. The support set represents the training data for that specific task, while the query set is used to evaluate the model's performance on that task. Following the \emph{N-way K-shot} paradigm, each support set consists of \emph{N} classes, each represented by \emph{K} image examples. In Meta DeepBDC, a BDC matrix is calculated for each support and query sample during each meta-training episode. The average of the support BDC matrices belonging to the same class forms the class prototype, and classification within each episode is performed by computing a similarity distribution of the query BDC matrix against all the class prototypes. Once the meta-fine-tuning is finalized, we evaluated its performance on various meta-testing tasks employing the same approach as in the meta-training phase.

In traditional meta-learning experiments, meta-training and meta-testing tasks typically involve unrelated sets of classes \cite{Snell17, Triantafillou19, Vinyals16}, e.g., distinguishing dog breeds in meta-training and cat breeds in meta-testing. However, some studies adopted a different approach by using distinct classes for meta-training and meta-testing that still belong to the same underlying data distribution. 
For instance, the authors in \cite{Jiang22, Singh21} used the same image modalities and anatomical structure but explored different types of diseases between meta-training and meta-testing.
We propose a novel approach wherein the model is tasked with similar classification tasks during meta-training and meta-testing but at different levels of granularity. Specifically, the meta-training classes ($C_{fine}$) constitute subgroups of the meta-testing ones ($C_{coarse}$), i.e., $|C_{fine}| > |C_{coarse}|$.
This design choice serves two purposes. First, during meta-training, the model is challenged to distinguish images at a finer granularity level. Second, since the number of \emph{ways} (distinct classes) remains constant between meta-training and meta-testing, and the total number of meta-training classes is higher, the episodes in meta-training become more diversely composed. These two facets amplify the complexity of the training phase, consequently strengthening the model's generalization capabilities. Meanwhile, since the model is tasked with similar but simpler tasks during meta-testing, we expect that its classification capabilities will improve accordingly.
Our reasoning behind this approach holds relevance in contexts where the meta-testing tasks carry higher clinical significance despite being easier. 

%Figure 3

\subsubsection{Fine-tuning ablations}

\noindent To evaluate the effectiveness of the described approach, we conducted three ablation fine-tuning experiments. To assess whether using finer-grained classes during meta-training compared to meta-testing contributes to higher model generalizability, we performed two additional meta-learning experiments by generating the meta-training set from a different source dataset than the one used for meta-testing. Formally, we considered an additional dataset $D^j$ where $j \neq i$ and new sets of fine-grained and coarse-grained classes, namely $C^j_{fine}$ and $C^j_{coarse}$ respectively. In the first ablation experiment, the meta-training classes belonged to $C^j_{fine}$, i.e., $D^j_{meta-train}=\{(x,y)_k, y_k \in C^j_{fine}\}$. In the second one, we considered the coarser-grained set of classes for meta-learning as well, $D^j_{meta-train}=\{(x,y)_k, y_k \in C^j_{coarse}\}$. Meta-validation and meta-testing sets remained unchanged.

Furthermore, as a third ablation experiment, we fine-tuned the pre-trained backbones using the classical fully-supervised approach. For this experiment, we utilized the same dataset splitting as in the proposed meta-learning approach but with a consistent class set across all subsets, denoted as $C^i_{coarse}$. Specifically, the datasets considered were as follows: $D^i_{train}=\{(x,y)_k, y_k \in C^i_{coarse}\}$, $D^i_{val}=\{(x,y)_k, y_k \in C^i_{coarse}\}$, and $D^i_{test}=\{(x,y)_k, y_k \in C^i_{coarse}\}$.

\subsection{Datasets}
\noindent We employed two datasets: the PI-CAI challenge dataset [dataset] \cite{Saha23}, consisting of 1500 mpMRI acquisitions of prostate cancer and the BreakHis dataset [dataset] \cite{Spanhol15}, specifically its processed version curated by Pereira [dataset] \cite{Pereira23}. Both datasets were confirmed to have been collected with institutional/ethical review board approval.

\subsubsection{PI-CAI dataset} 
\noindent For our experiments, we focused exclusively on T2-weighted images of patients with both cancerous and benign lesions. All acquisitions of benign lesions, specifically those without biopsy or with a negative biopsy (for which we lacked ground truth), were used as $D_{unlab}$. This dataset comprised 11202 images from 849 patients. Conversely, we utilized the acquisitions of cancerous lesions as $D_{lab}$ for the fine-tuning phase. The labeled dataset contains $2049$ images from $382$ patients, which we divided into $1611$ for training, $200$ for validation, and $238$ for testing. During the splitting process, we ensured patient stratification, i.e., all images from the same patient were grouped in the same subset, avoiding any data leakage.

\begin{figure*}[ht]
 \subfloat[PI-CAI dataset.]{\includegraphics[width=\linewidth]{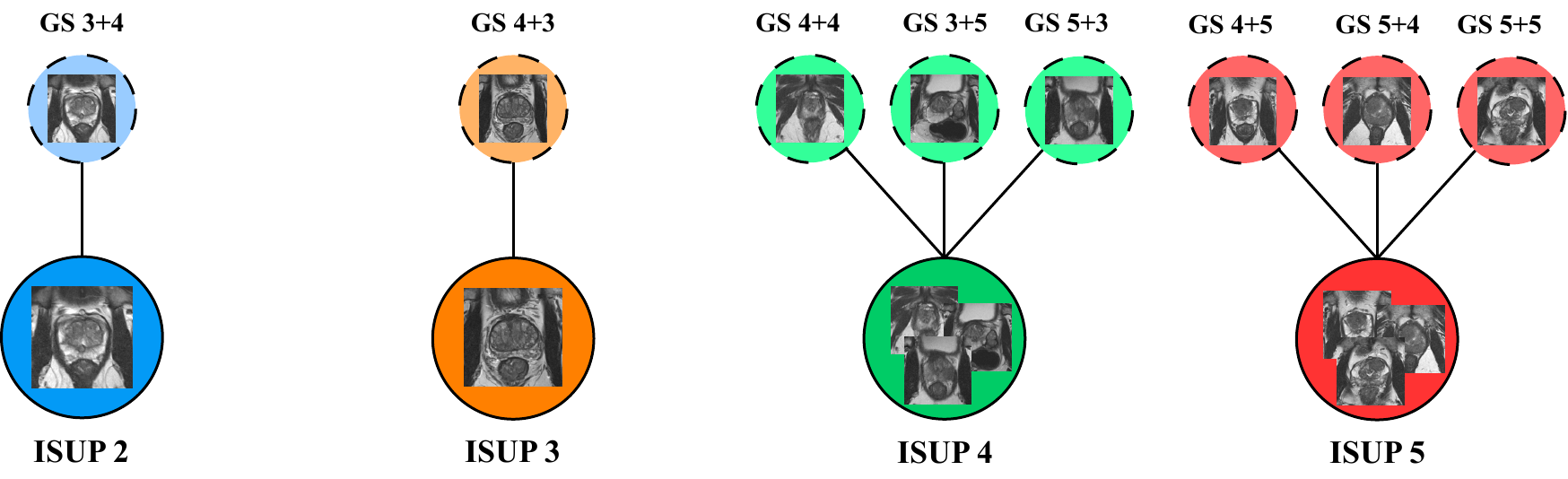}
\label{fig:grouped_labels_picai}}
\hfill
%Figure 2b
\subfloat[BreakHis dataset.]{\includegraphics[width=\linewidth]{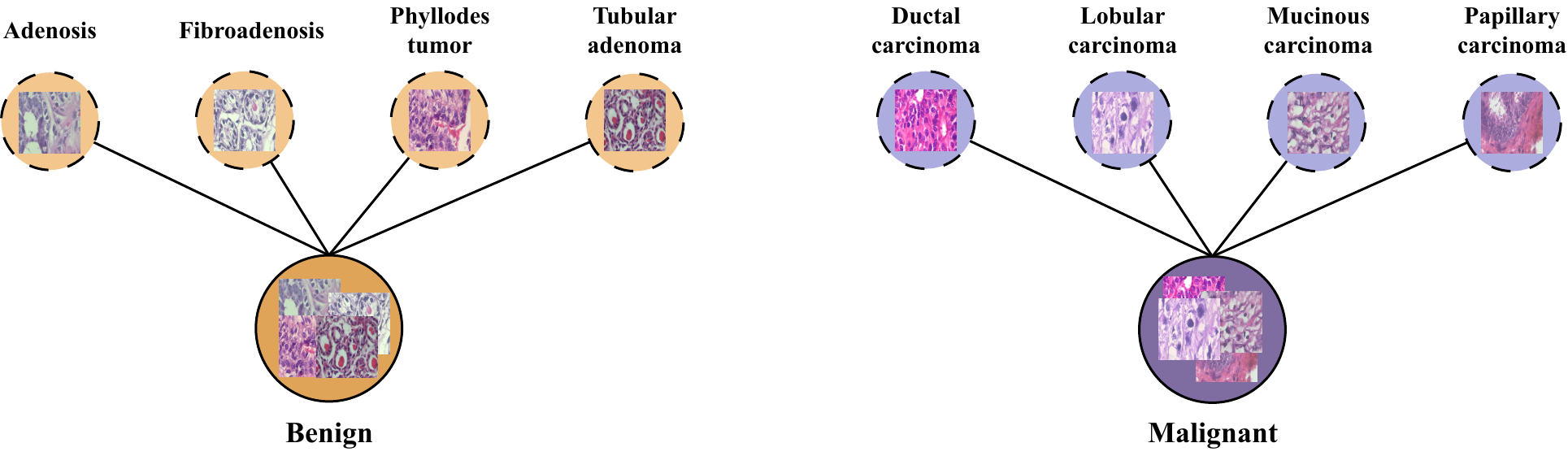}
\label{fig:grouped_labels_breakhis}}
\caption{Relationship between fine and coarse labels in (a) PI-CAI dataset and (b) BreakHis dataset.}
\end{figure*}

Each mpMRI acquisition was provided together with a biopsy report indicating the severity of each lesion, assessed by the Gleason Score (GS), which can assume values ranging from $1+1$ to $5+5$ based on the severity of the two most common patterns in the biopsy specimen. 
Related to the GS, each lesion was assigned a prognostic score according to the ISUP guidelines \cite{Egevad16}, which ranges from $1$ to $5$, according to the tumor severity.  
Fig. \ref{fig:grouped_labels_picai} visually represents the relationship between ISUP and GS. For this study, our focus was exclusively on lesions with ISUP scores $\geq 2$, as annotations for ISUP-$1$ lesions, necessary to perform the image pre-processing steps, were unavailable. Consequently, we considered the eight GS classes (from $3+4$ to $5+5$) as the $C_{fine}$ class set and the four ISUP classes (from $2$ to $5$) as the $C_{coarse}$ set. For clarification purposes, we provided the labeling criteria for the training, validation, and test subsets in Table \ref{tab:data_labeling}.

%Figure 2a

Furthermore, to emulate a realistic scenario with distinct distributions in training and testing data, we considered only Siemens vendor data for training and only Philips vendor data for both validation and testing. We opted for the analogous distributions among validation and test datasets since, as emphasized by Setlur \emph{et al.} \cite{Setlur21}, using validation samples that are not independent and identically distributed with the test samples can result in unreliable results when determining the optimal model, especially the one that maximizes performance on the test set. The vendor used for the training dataset was also employed for the pre-training dataset, ensuring a complete separation in data distributions between the pre-training/training and evaluation phases.

As for the data pre-processing, we utilized the provided whole prostate segmentation to extract the mask centroid for each slice. We standardized the field of view (FOV) at $100$ mm %in both $x$ ($FOV_x$) and $y$ ($FOV_y$)  directions 
to ensure consistency across all acquisitions and cropped each image based on this value around the found centroid. We determined the number of rows ($N_{rows}$) and columns ($N_{cols}$) for the fixed FOV, leveraging the pixel spacing in millimeters along the x-axis ($px$) and the y-axis ($py$) according to the following formulae: $N_{cols} = \frac{FOV_x}{px}$ and $N_{rows} = \frac{FOV_y}{py}$.
Furthermore, we resized all images to a uniform matrix size of $128 \times 128$ pixels to maintain a consistent pixel count. Finally, we performed a z-score normalization on all images using an in-volume approach. 

\subsubsection{BreakHis dataset}

\noindent The dataset consists of 7909 microscopic images of breast tumor tissues from 82 patients, taken at four different magnification levels: 40X, 100X, 200X, and 400X. Each image has a size of 224 by 224 pixels. We used the images at 40X, 100X, and 200X magnification levels for a total of 6090 images as $D_{unlab}$ by ignoring their labels. On the other hand, we used the images at 400X magnification level, namely 1819 images, as $D_{lab}$ to perform the fine-tuning step. We split this dataset into training, validation, and test sets, with  1475, 165, and 183 images, respectively. The dataset has eight classes of lesions which we considered as $C_{fine}$ set: \emph{adenosis}, \emph{tubular adenoma}, \emph{fibroadenoma}, \emph{phyllodes tumor}, \emph{papillary carcinoma}, \emph{lobular carcinoma}, \emph{mucinous carcinoma}, and \emph{ductal carcinoma}. The dataset also provides a binary classification of the lesions into benign and malignant, which we considered as $C_{coarse}$ set. Specifically, the first four fine classes are benign, while the last four are malignant. We provided a visual representation of the relatioship between fine and coarse classes in Fig. \ref{fig:grouped_labels_breakhis}. For clarification purposes, we also provided the labeling criteria for training, validation, and test sets of the BreakHis dataset in Table \ref{tab:data_labeling}.

\begin{table*}[ht]
    \centering
    \caption{A summary of the labeling procedure for the two examined datasets. ISUP = International Society of Urological Pathology, GS = Gleason Score, AD = Adenosis, TA = Tubular Adenoma, FA = Fibroadenoma, PT = Phyllodes Tumor, PC = Papillary Carcinoma, LC = Lobular Carcinoma, MC = Mucinous Carcinoma, DC = Ductal Carcinoma.}
    \label{tab:data_labeling}
    \resizebox{\textwidth}{!}{
    \begin{tabular}{|c|c|c|c|}
    \hline
        \textbf{Dataset} & \textbf{Subset} &\textbf{Meta-fine-tuning labels} & \textbf{Fully-supervised fine-tuning labels}\\
        \hline
         \multirow{3}{*}{PI-CAI} & (Meta) Training & \makecell{GS: $3+4$, $4+3$, $4+4$, $3+5$, $5+3$, $4+5$, $5+4$, $5+5$} & ISUP: $2$, $3$, $4$, $5$\\
         \cline{2-4}
         & (Meta) Validation & ISUP: $2$, $3$, $4$, $5$ & ISUP: $2$, $3$, $4$, $5$\\
         \cline{2-4}
        & (Meta) Test &  ISUP: $2$, $3$, $4$, $5$ & ISUP: $2$, $3$, $4$, $5$\\
        \hline 
         \multirow{3}{*}{BreakHis} & (Meta) Training & \makecell{AD, TA, FA, PT, PC, LC, MC, DC} & Benign, Malignant\\
         \cline{2-4}
         & (Meta) Validation & Benign, Malignant & Benign, Malignant\\
         \cline{2-4}
        & (Meta) Test & Benign, Malignant & Benign, Malignant\\
    \hline
    \end{tabular}}
\end{table*}

%%%%%%%%%%%%%%%%%%%%%%%%%%%%%%%%%%%%%%%%%%%

\subsection{Experiments}
\noindent We performed all our experiments using four popular CNNs as backbones, namely, ResNet-18, ResNet-50, VGG-16 and DenseNet-121.

\subsubsection{Pre-training}

\noindent In both the vanilla SSL and SSL+IP-IRM  pre-training phases, we conducted a hyperparameter optimization to determine the best configuration for our experiments. This involved evaluating various values for weight decay (WD), batch size (BS), and the number of epochs, as outlined in Table \ref{tab:hyperparameters}. Following the approach established in the original SimCLR paper \cite{Chen20}, we set the learning rate (LR) as a function of the BS, as follows:

\begin{align}
    LR = \frac{0.3\cdot BS}{256}.
\end{align}

\noindent We opted for large BS values, aligning with the observation made by Chen \emph{et al.} \cite{Chen20} that larger BS generally yield improved performance. Consistent with the original implementation, we conducted our experiments using the SGD optimizer and employed the cosine decay schedule to adjust the learning rate throughout epochs. To mitigate the risk of performance degradation associated with inconsistent input sizes \cite{Zhang23}, we maintained the same input size during both the pre-training and fine-tuning phases.

As highlighted in \cite{Zhang23}, performance on proxy tasks does not consistently correlate positively with performance on downstream tasks. Therefore, we selected the optimal pre-training hyperparameters as the ones that maximize the performance of the fine-tuned model. Finally, regarding the composition of augmentations, we utilized the combination that yielded the best performance according to the original paper, i.e., random cropping and random color distortion augmentations.

\begin{table*}[ht]
    \centering
     \caption{A summary of the hyperparameters optimized during pre-training and meta-fine-tuning phases. The provided learning rate values must be considered as initial values before decay. The learning rate is decreased by a factor of 10 with each decay epoch.}
    \label{tab:hyperparameters}
    \begin{tabular}{|c|c|}
    %\begin{tabular}{\linewidth}
    \hline
         \textbf{Hyperparameter} & \textbf{Values}\\
         \hline
         \multicolumn{2}{|c|}{Pre-training phase} \\
        \hline
         Weight decay & [$10^{-2}$, $10^{-3}$, $10^{-4}$, $10^{-5}$] \\
         Batch size & [128,256,512]\\
         Epochs & [100,400]\\ \hline 
         \multicolumn{2}{|c|}{Meta-fine-tuning phase} \\
         \hline
         Learning rate & [$10^{-1}$, $10^{-2}$, $10^{-3}$, $10^{-4}$]\\
         Weight decay & [$10^{-2}$, $10^{-3}$, $10^{-4}$, $10^{-5}$] \\
         Epochs & [100,400] \\
         Decay epochs for 100 epochs & [20,50]\\
         Decay epochs for 400 epochs & [80,200]\\ \hline
    \end{tabular}
\end{table*}

\subsubsection{Fine-tuning}

\noindent To establish a fair comparison across all experiments, we defined a baseline model, namely the fully-supervised-pretrained model on the ImageNet dataset, fine-tuned using the traditional fully-supervised approach. We optimized its hyperparameters, such as LR, WD, and the number of epochs, and we employed these optimized hyperparameters for all subsequent fine-tuning experiments. To enhance the training process, particularly in terms of optimization and generalization, we exploited a learning rate decay strategy. This strategy involves reducing the LR by a factor of 10 after a predetermined number of warm-up epochs. We reported the hyperparameters evaluated during the baseline optimization in Table \ref{tab:hyperparameters}.

As for our meta-fine-tuning experiments, we performed a \emph{4-way K-shot} training for the PI-CAI dataset and a \emph{2-way K-shot} for the BreakHis dataset. For both datasets, we evaluated both \emph{1-shot} and \emph{5-shot} configurations, resulting in support sets containing either one or five examples per class, respectively.  
In contrast, the query set always consisted of 10 examples for each class. We conducted 600 episodes throughout both the meta-training and meta-validation phases, as well as the meta-testing phase. Additionally, we replicated the meta-testing evaluation five times.

In all fine-tuning experiments, we refrained from using any data augmentation. This decision was based on evidence suggesting that data augmentation can diminish the positive impact of SSL pre-training and may even negatively impact model performance, as discussed in \cite{Zhang23}.

\subsubsection{Evaluation metrics and loss function}
\noindent To evaluate the performance of our models, we employed both binary and multi-class Area Under the ROC Curve (AUROC), which proves more stable in the context of accuracy when handling the imbalance often found in medical imaging datasets. For the computation of multi-class AUROC, we adopted the \emph{One-vs-rest} approach, averaging the binary AUROC between each class and all the others. In the meta-learning experiments, AUROC was assessed in each episode, and the final AUROC performance was determined by averaging the AUROC of each episode during the meta-testing phase.

Given our objective of maximizing the AUROC metric, we adopted the AUC margin loss (AUC-M) as our loss function. Introduced in \cite{Yuan21} and defined in the LibAUC library \cite{Yuan23}, AUC-M loss is a surrogate loss function designed to maximize the ROC curve. Specifically, AUC-M loss is a min-max loss function that encourages the model to learn a decision boundary that separates positive and negative examples with a large margin. This property makes the AUC-M loss more robust to noisy data and not adversely affected by simple data.
As an optimizer algorithm, following \cite{Yuan23}, we employed the Proximal Epoch Stochastic method (PESG) \cite{Guo23}, which is a stochastic method designed to solve smooth non-convex strongly-concave min-max problems, such as deep AUC maximization \cite{Guo23}.

%%%%%%%%%%%%%%%%%%%%%%%%%%%%%%%%%%%%%%%%%%%%%%%%%%%%%%%%%%%%%

\section{Results}
\label{sec:results}

\noindent The hyperparameter optimization of the baseline model led to the following values for both ResNet-18 and ResNet-50: $10^{-2}$ LR, $10^{-2}$ WD, and $100$ epochs for the PI-CAI dataset, and $10^{-1}$ LR, $10^{-2}$ WD, and $100$ epochs for the BreakHis dataset. For the VGG-16, we found that the optimal hyperparameters are $10^{-3}$ LR, $10^{-5}$ WD and $100$ epoch for the PI-CAI dataset, $10^{-3}$ LR, $10^{-4}$ WD and $100$ epochs for the BreakHis dataset. Finally, concerning the DenseNet-121 backbone, the hyperparameter optimization provided the following optimal hyperparameters: $10^{-2} $ LR, $10^{-4}$ WD and $100$ epochs for the PI-CAI dataset and $10^{-2}$ LR, $10^{-5}$ WD and $100$ epochs for the BreakHis dataset.

\begin{figure*}
 \subfloat[PI-CAI dataset.]{\includegraphics[width=\linewidth]{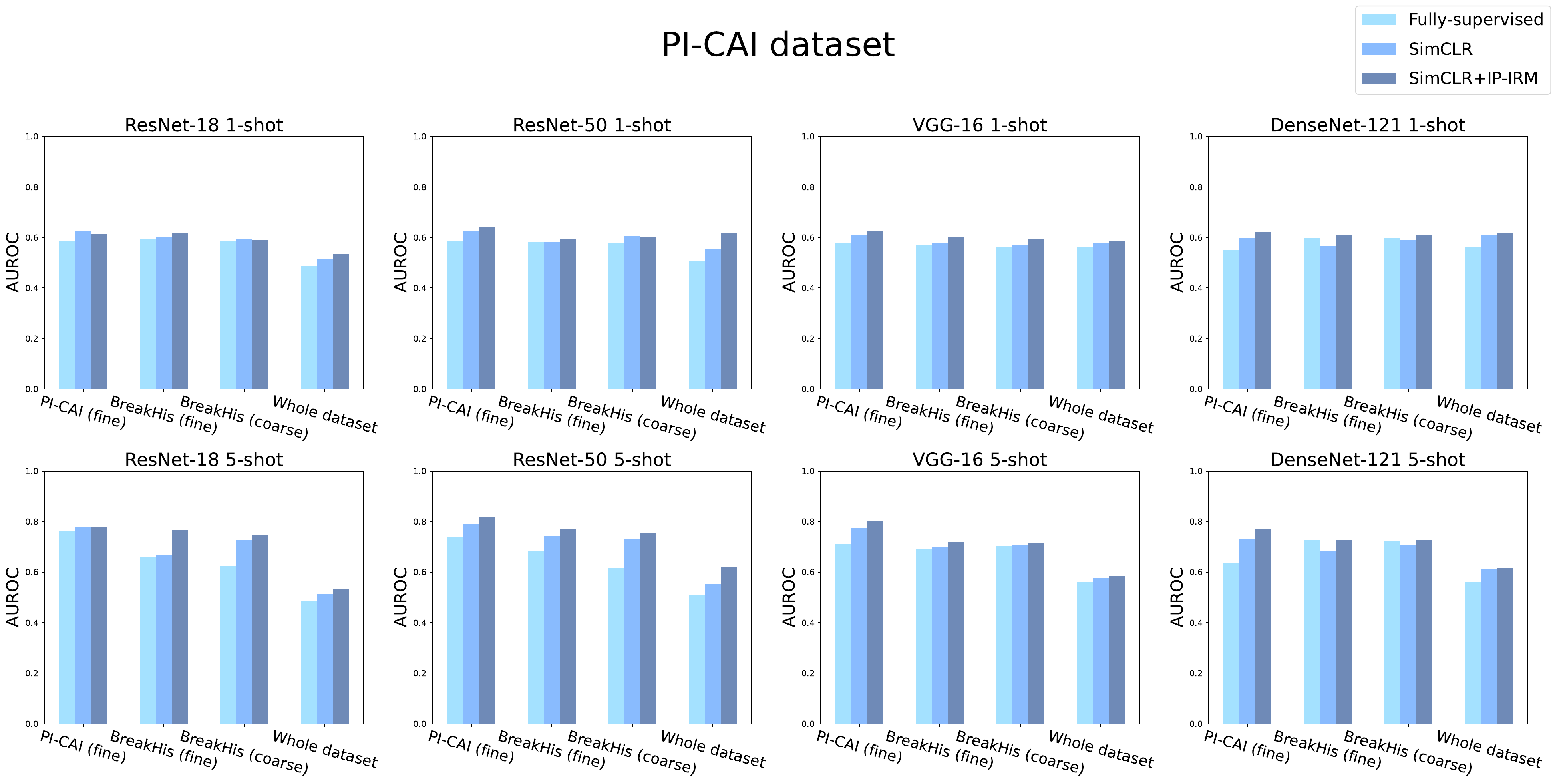}
\label{fig:bar_chart_picai}}
\hfill
%Figure 3b
\subfloat[BreakHis dataset.]{\includegraphics[width=\linewidth]{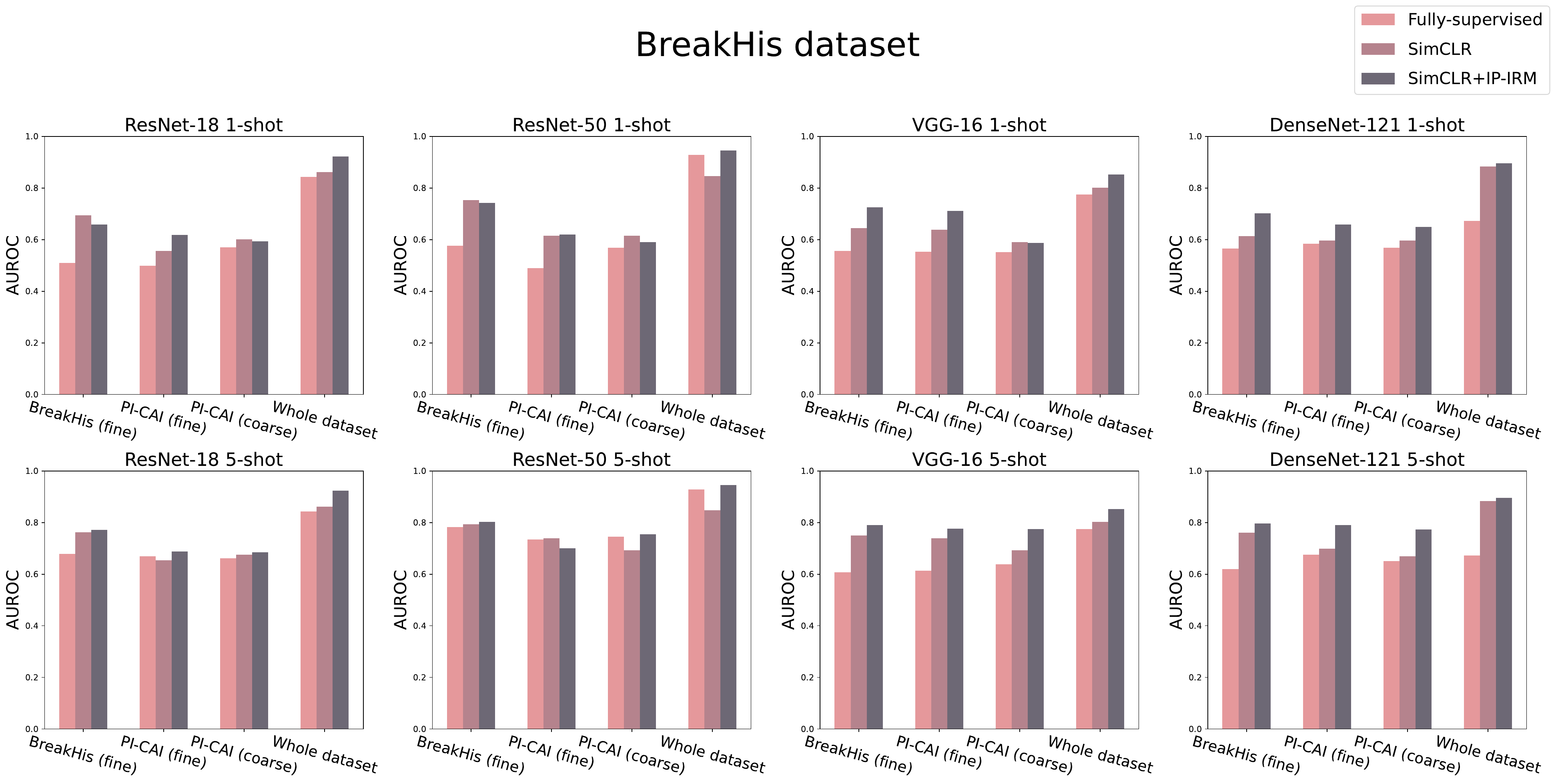}
\label{fig:bar_chart_breakhis}}
\caption{Results visual representation for (a) PI-CAI dataset and (b) BreakHis dataset. Each plot represents the results of a backbone in a 1-shot or 5-shot setting. We represented each fine-tuning scheme with a three-column group. In each group, the three colours indicate the three pre-training approaches.}
\end{figure*}

Concerning model performance, we presented the results for the PI-CAI dataset in Table \ref{tab:results_picai}. There are two main perspectives from which to analyze the results. The first perspective pertains to the pre-training type, which varies according to the Table \ref{tab:results_picai} rows. Specifically, for each backbone, the last row (\textbf{SimCLR+IP-IRM}) indicates the pre-training step of the proposed approach, while the other two rows (\textbf{Fully-supervised} and \textbf{SimCLR}) represent the ablation pre-training experiments. The second perspective relates to the fine-tuning method, which varies with the columns in Table \ref{tab:results_picai}. We presented the results of the proposed meta-fine-tuning approach in the grouped column \textbf{Meta-train on PI-CAI \emph{(fine)}}. In contrast, we detailed the ablation fine-tuning experiments in the remaining three columns, namely \textbf{Meta-train on BreakHis \emph{(fine)}}, \textbf{Meta-train on BreakHis \emph{(coarse)}}, and \textbf{Fully-supervised}.
The \textbf{Meta-train on BreakHis \emph{(fine)}} and \textbf{Meta-train on BreakHis \emph{(coarse)}} columns showcase the results wherein we utilized different sources for the meta-train and meta-testing datasets, considering both fine-grained and coarse-grained classes during meta-training, respectively. On the other hand, the \textbf{Fully-supervised} column presents the results of a classical fully-supervised fine-tuning on the entire dataset.
To offer a more immediate visualization, we depicted the results for the PI-CAI dataset in a bar chart presented in Fig. \ref{fig:bar_chart_picai}. In this representation, for each backbone and each k-shot setting (1-shot or 5-shot), we delineated the outcomes of various pre-training experiments using bars of different colors, and we portrayed the distinct fine-tuning experiments through separate groups of bars.

Similarly, we presented the results on the BreakHis dataset in Table \ref{tab:results_breakhis}. As in the approach taken for the PI-CAI datasets, we organized the results of different pre-training experiments as distinct rows and the fine-tuning experiments as separate columns in Table \ref{tab:results_breakhis}. In this context, we detailed the outcomes of the proposed fine-tuning approach in column \textbf{Meta-train of BreakHis \emph{(fine)}}. Conversely, we delineated the ablation experiment in columns  \textbf{Meta-train on PI-CAI \emph{(fine)}}, \textbf{Meta-train on PI-CAI \emph{(coarse)}}, where a diverse source dataset for meta-training and meta-testing is utilized, and in column \textbf{Fully-supervised}, representing the conventional fully-supervised fine-tuning.
In line with the PI-CAI dataset, we illustrated the results in Fig. \ref{fig:bar_chart_breakhis}. In the same way as for the PI-CAI dataset, for each backbone and k-shot setting we represented the diverse pre-training experiments through bars of different colors while differentiating the fine-tuning experiments by distinct bar groups.

\begin{table*}[ht]
    \centering
    \caption{Test set results in terms of multi-class AUROC for the PI-CAI dataset. In all the experiments, the downstream task consists of classifying the images into four ISUP classes. We experimented with performing meta-training on both finer-grained classes (eight GS classes and eight breast tumor classes, denoted as \emph{fine}) and coarser ones (four malignant breast tumor classes, denoted as \emph{coarse}). For each experiment, we highlighted the proposed and ablation approaches accordingly.}
    \label{tab:results_picai}
    \resizebox{\textwidth}{!}{
    \begin{tabular}{|c|c|c|c|c|c|c|c|c|}

        \hline
         \textbf{Backbone} & \textbf{Pre-training type} & \multicolumn{7}{|c|}{\textbf{Fine-tuning type}} \\ \hline
        & & \multicolumn{2}{|c|}{\makecell{\textbf{Meta-train on PI-CAI (\emph{fine})}\\ \textbf{(Proposed)}}} &  \multicolumn{2}{|c|}{\makecell{\textbf{Meta-train on BreakHis (\emph{fine})}\\ \textbf{(Ablation)}}} & \multicolumn{2}{|c|}{\makecell{\textbf{Meta-train on BreakHis (\emph{coarse})}\\ \textbf{(Ablation)}}}& \makecell{\textbf{Fully-supervised}\\ \textbf{(Ablation)}}\\ \hline
        
           & & \textbf{4-way 1-shot} &\textbf{4-way 5-shot} & \textbf{4-way 1-shot} &\textbf{4-way 5-shot} & \textbf{4-way 1-shot} &\textbf{4-way 5-shot}& \textbf{Whole dataset}\\ \hline
         
         \multirow{3}{*}{ResNet-18} & \textbf{Fully-supervised (Ablation)}  & 0.585 & 0.763 & 0.594 & 0.659 & 0.588 & 0.626 & 0.487\\
                                   & \textbf{SimCLR (Ablation)} & \textbf{0.624} & 0.779  & 0.600 & 0.667 & \textbf{0.592} & 0.727 & 0.515\\
                                   & \textbf{SimCLR + IP-IRM (Proposed)}  & 0.615 & \textbf{0.780} & \textbf{0.618} & \textbf{0.767} &0.591 &\textbf{0.749} & \textbf{0.533}\\
         \hline
         \multirow{3}{*}{ResNet-50} & \textbf{Fully-supervised (Ablation)}  & 0.587 & 0.740 & 0.582 & 0.683 &0.578 &0.616& 0.509\\
                                   & \textbf{SimCLR (Ablation)}& 0.628 & 0.790 & 0.581 & 0.744 & \textbf{0.605}& 0.732 & 0.553\\
                                   & \textbf{SimCLR + IP-IRM (Proposed)} & \textbf{0.640} & \textbf{0.821} & \textbf{0.595} & \textbf{0.773} & 0.602&\textbf{0.756} & \textbf{0.620} \\\hline
        \multirow{3}{*}{VGG-16} & \textbf{Fully-supervised (Ablation)} & 0.579 & 0.712 & 0.569 & 0.693 & 0.563 & 0.704 & 0.562 \\
                                   & \textbf{SimCLR (Ablation)}& 0.608 &  0.776 & 0.578 & 0.701 & 0.571& 0.706 & 0.576\\
                                   & \textbf{SimCLR + IP-IRM (Proposed)} & \textbf{0.626} & \textbf{0.803} & \textbf{0.603} & \textbf{0.720}&\textbf{0.593} & \textbf{0.717} & \textbf{0.585} \\\hline
        \multirow{3}{*}{DenseNet-121} & \textbf{Fully-supervised (Ablation)} & 0.549 & 0.635 & 0.597 & 0.727 & 0.599 & 0.726 & 0.560 \\
                                   & \textbf{SimCLR (Ablation)}& 0.597 &  0.730& 0.566 & 0.685 & 0.589 & 0.709 & 0.611\\
                                   & \textbf{SimCLR + IP-IRM (Proposed)} & \textbf{0.621} & \textbf{0.771} &  \textbf{0.612} & \textbf{0.729}& \textbf{0.610} & \textbf{0.727} & \textbf{0.617} \\\hline
    \end{tabular}}
\end{table*}

\begin{table*}[ht]
    \centering
     \caption{Test set results in terms of binary AUROC for the BreakHis dataset. In all the experiments, the downstream task consists of classifying the images into two classes (Benign vs. Malignant). We experimented with performing meta-training on both finer-grained classes (eight breast tumor classes and eight GS classes, denoted as \emph{fine}) and coarser ones (four ISUP classes, denoted as \emph{coarse}). For each experiment, we highlighted the proposed and ablation approaches accordingly.}
    \label{tab:results_breakhis}
    \resizebox{\textwidth}{!}{
    \begin{tabular}{|c|c|c|c|c|c|c|c|c|}
    \hline
         \textbf{Backbone} & \textbf{Pre-training type} & \multicolumn{7}{|c|}{\textbf{Fine-tuning type}} \\ \hline
        & & \multicolumn{2}{|c|}{\makecell{\textbf{Meta-train on BreakHis (\emph{fine})} \\ \textbf{(Proposed)}}} & \multicolumn{2}{|c|}{\makecell{\textbf{Meta-train on PI-CAI (\emph{fine})}\\ \textbf{(Ablation)}}} & \multicolumn{2}{|c|}{\makecell{\textbf{Meta-train on PI-CAI (\emph{coarse})}\\ \textbf{(Ablation)}}}& \makecell{\textbf{Fully-supervised}\\ \textbf{(Ablation)}}\\ \hline
        
           & & \textbf{2-way 1-shot} &\textbf{2-way 5-shot} & \textbf{2-way 1-shot} &\textbf{2-way 5-shot} & \textbf{2-way 1-shot} &\textbf{2-way 5-shot}&\textbf{Whole dataset}\\ \hline
         
         \multirow{3}{*}{ResNet-18} & \textbf{Fully-supervised (Ablation)}  & 0.510 & 0.678 & 0.499&0.669 & 0.570 & 0.661 & 0.843\\
                                   & \textbf{SimCLR (Ablation)} & \textbf{0.694} &  0.763 &0.556 & 0.654 &\textbf{0.601} & 0.675& 0.862\\
                                   & \textbf{SimCLR + IP-IRM (Proposed)} & 0.658 & \textbf{0.772} &\textbf{0.618} & \textbf{0.688}& 0.594& \textbf{0.685} &  \textbf{0.923}\\
         \hline
         \multirow{3}{*}{ResNet-50} & \textbf{Fully-supervised (Ablation)}  & 0.576 & 0.783 & 0.490 & 0.735& 0.568& 0.746 & 0.928\\
                                   & \textbf{SimCLR (Ablation)}& \textbf{0.754} & 0.793 &0.615 &\textbf{0.739} &\textbf{0.615} & 0.692 & 0.847 \\
                                   & \textbf{SimCLR + IP-IRM (Proposed)} & 0.742 & \textbf{0.802} & \textbf{0.620}& 0.700 &0.591 & \textbf{0.754} &  \textbf{0.946} \\ \hline
        \multirow{3}{*}{VGG-16} & \textbf{Fully-supervised (Ablation)}  & 0.556 & 0.608 & 0.554 & 0.613 & 0.551 & 0.639 & 0.775 \\
                                   & \textbf{SimCLR (Ablation)}& 0.644 &   0.750 & 0.638 & 0.739 &\textbf{0.590} & 0.692 & 0.802\\
                                   & \textbf{SimCLR + IP-IRM (Proposed)} & \textbf{0.725} & \textbf{0.791} & \textbf{0.712}& \textbf{0.777} & 0.588& \textbf{0.775} &  \textbf{0.853}\\\hline
        \multirow{3}{*}{DenseNet-121} & \textbf{Fully-supervised (Ablation)} & 0.565 & 0.620& 0.584 & 0.675 & 0.568 & 0.651 &  0.673\\
                                   & \textbf{SimCLR (Ablation)}& 0.613 & 0.761 & 0.597 & 0.699 & 0.596& 0.669 & 0.883\\
                                   & \textbf{SimCLR + IP-IRM (Proposed)} & \textbf{0.702} & \textbf{0.797} & \textbf{0.658} & \textbf{0.790} & \textbf{0.649} & \textbf{0.774} & \textbf{0.896}\\\hline
    
    \end{tabular}}
\end{table*}

%Figure 3a

\noindent \textbf{Pre-training results.} For each fine-tuning experiment, in both Table \ref{tab:results_picai} and Table \ref{tab:results_breakhis}, we highlighted the pre-training approach that performs best in bold. 
Specifically, for the PI-CAI dataset, SimCLR+IP-IRM pre-training produces the most favorable results for all the 5-shot and fully-supervised fine-tuning experiments. In contrast, for three out of six 1-shot experiments (two for the ResNet-18 and one for the ResNet-50), the optimal performance is achieved through SimCLR alone. A similar trend is evident in the BreakHis dataset, where SimCLR alone serves as the best pre-training step for five 1-shot configurations (two for the ResNet-18, two for the ResNet-50 and one for the VGG-16). However, for all 5-shot configurations except one, as well as the fully-supervised approach, consistent enhancements are observed with SimCLR+IP-IRM pre-training.

\noindent \textbf{Meta-fine-tuning results.} In describing the meta-fine-tuning results, among the pre-training approaches (three for each meta-fine-tuning strategy), we refer to the one that provided the best AUROC value. Concerning the PI-CAI dataset, according to Table \ref{tab:results_picai}, the optimal performance is obtained by the proposed approach (first fine-tuning type column of Table \ref{tab:results_picai}), yielding 0.780 AUROC for ResNet-18, 0.821 for ResNet-50, 0.803 for VGG-16 and 0.771 for DenseNet-121. In terms of the ablation experiments, employing the meta-learning approach with granularity and dataset shift (second fine-tuning type column in Table \ref{tab:results_picai}) provides the second-best performance, achieving the highest AUROC of 0.767 for ResNet-18, 0.773 for ResNet-50, 0.720 for VGG-16 and 0.729 for DenseNet-121. Lastly, when meta-training is conducted on coarser classes (third fine-tuning type column in Table \ref{tab:results_picai}), classification performance is further degraded, with an AUROC of 0.749 for ResNet-18, 0.756 for ResNet-50, 0.717 for VGG-16 and of 0.727 for DenseNet-121.

Exploring the BreakHis dataset, the proposed meta-fine-tuning configuration (first fine-tuning type column of Table \ref{tab:results_breakhis}) yields the most favorable results for all the backbones, with AUROCs of 0.772, 0.802, 0.791 and 0.797 for ResNet-18, ResNet-50, VGG-16 and DenseNet-121, respectively. Considering the ablation experiments, for ResNet-18, VGG-16 and DenseNet-121, the second-best performance is observed when both granularity and dataset source shift are present (second fine-tuning type column of Table \ref{tab:results_breakhis}), resulting in 0.688, 0.777 and 0.790 AUROC, respectively. On the other hand, for the same three backbones, the least favorable outcome is obtained when coarser-grained classes are employed for meta-training (third fine-tuning type column of Table \ref{tab:results_breakhis}), yielding an AUROC of 0.685, 0.775 and 0.774, respectively. Conversely, for ResNet-50, using coarser classes for meta-training produces superior results, achieving a 0.754 AUROC compared to 0.739 AUROC obtained with finer-grained classes.

\noindent \textbf{Fully-supervised fine-tuning results.}
As the last ablation experiment, we considered performing a fully-supervised fine-tuning on the entire dataset. For the PI-CAI dataset, this fine-tuning strategy (results provided in the fourth fine-tuning type column of Table \ref{tab:results_picai}) exhibits the worst performance w.r.t. all the 5-shot settings and even w.r.t. several 1-shot, especially when the proposed meta-fine-tuning approach is employed. Indeed, for the four backbones the higher AUROC is 0.533, 0.620, 0.585 and 0.617, respectively. This underscores that the meta-learning approach significantly enhances the generalization ability of both backbones, no matter the low-data regime, and that the proposed meta-fine-tuning approach surpasses the fully-supervised one even in the 1-shot regime. Conversely, for the BreakHis dataset, the fully-supervised fine-tuning approach (fourth fine-tuning type column of Table \ref{tab:results_breakhis}) exhibits considerable strength, overcoming all the results provided by the meta-fine-tuning strategies. Indeed, this approach yields 0.923, 0.946, 0.853 and 0.896 AUROC for ResNet-18, ResNet-50, VGG-16 and DenseNet-121, respectively.

%%%%%%%%%%%%%%%%%%%%%%%%%%%%%%%%%%%%%%%%%%%%%%%%%%%%%%%%%%%%%

 \section{Discussion}
\label{sec:discussion}

\noindent In this work, we introduced an approach to enhance the performance of models trained in an FSL paradigm. The proposed strategy comprises a pre-training step that utilizes SSL coupled with the IP-IRM algorithm, which forces the feature disentanglement, ultimately guiding the model in learning more robust feature representation for the downstream task and making it distinguish between invariant and task-specific features. In addition, we proposed an enhanced meta-learning framework that leverages classes between the meta-training and meta-testing phases at different granularity levels. This approach aims to improve the model's generalization ability by exposing it to a more diverse and structured meta-training phase.

To evaluate the efficacy of the proposed pre-training strategy, we conducted two additional ablation experiments, i.e., a vanilla SSL method and a classical fully-supervised approach. As for the pre-training phase, to assess the effectiveness of the proposed meta-learning framework, we explored three additional ablation fine-tuning experiments as well, namely two meta-fine-tuning and a fully-supervised fine-tuning experiment. We delve into the meta-fine-tuning experiments in the following:
\begin{itemize}
    \item \textbf{Same source dataset with granularity shift (Proposed)}: Both meta-training and meta-testing sets derive from the same source dataset, but the granularity of classes differs. We employed finer-grained classes for meta-training to promote detailed feature learning and coarser-grained ones for meta-testing to assess the model's ability to generalize to broader categories.
    \item \textbf{Different source datasets with granularity shift (Ablation)}: Meta-training and meta-testing sets are drawn from distinct source datasets, but the granularity relationship between the classes remains consistent as in the previous case. This approach aims to highlight the effectiveness of the proposed granularity shift even when meta-training and meta-testing have different source datasets.
    \item \textbf{Different Source Datasets with granularity consistency (Ablation)}: Meta-training and meta-testing sets stem from different source datasets, but the granularity between classes is maintained consistently. We designed this approach to evaluate whether reducing class granularity during the meta-training impacts the model's generalizability in the subsequent meta-testing phase.
\end{itemize}

In terms of pre-training, concerning the PI-CAI dataset, the results presented in Table \ref{tab:results_picai} demonstrate that using the IP-IRM algorithm consistently provides the best-performing outcomes across all fully-supervised and 5-shot fine-tuning experiments. However, in the 1-shot experiments, the SimCLR alone outperforms the SimCLR+IP-IRM in three out of twelve configurations.
Regarding the BreakHis dataset, the proposed pre-training approach exhibits consistent improvement in most of the experiments. However, similar to the PI-CAI dataset, some 1-shot experiments reveal that SimCLR alone yields better results w.r.t. the IP-IRM algorithm. One possible explanation for this behaviour is that during the pre-training phase, the IP-IRM algorithm may lead to an over-disentangled data representation, resulting in excessively specialized representations and overly sensitive to variations in the training set. Consequently, in a 1-shot setting where the model must learn from only one example and may not have the opportunity to grasp all relevant features, it might struggle to generalize effectively, thus providing worse performance.

\begin{figure*}[ht]
\centering
\subfloat[ResNet-18, PI-CAI dataset]{\includegraphics[width=0.25\textwidth]{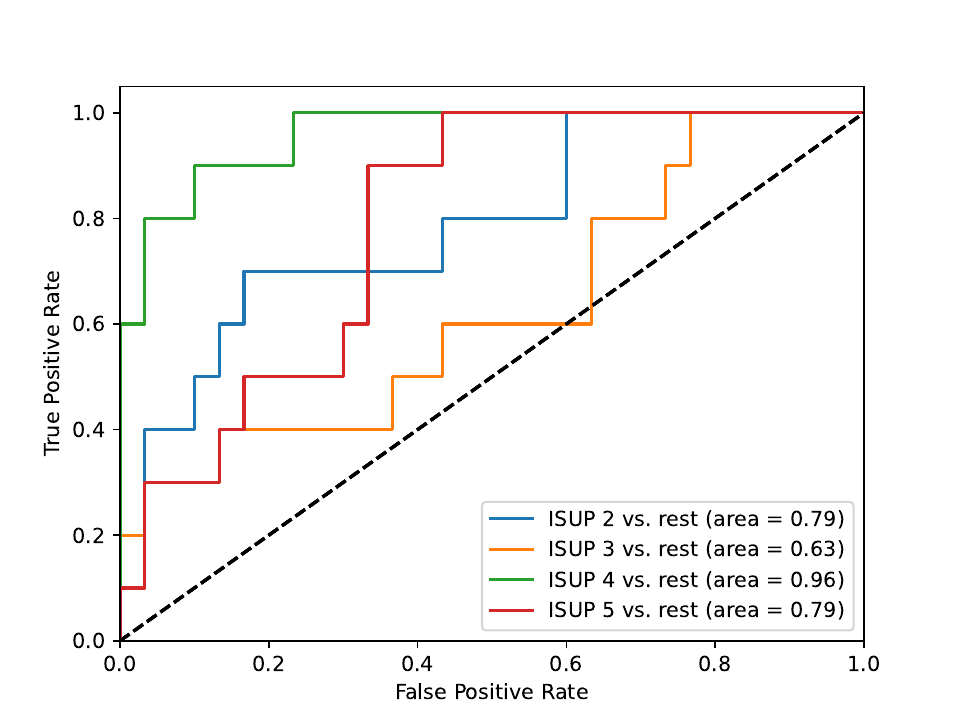}
\label{fig:picai_resnet18}}
%\hfil
\subfloat[ResNet-50, PI-CAI dataset]{\includegraphics[width=0.25\textwidth]{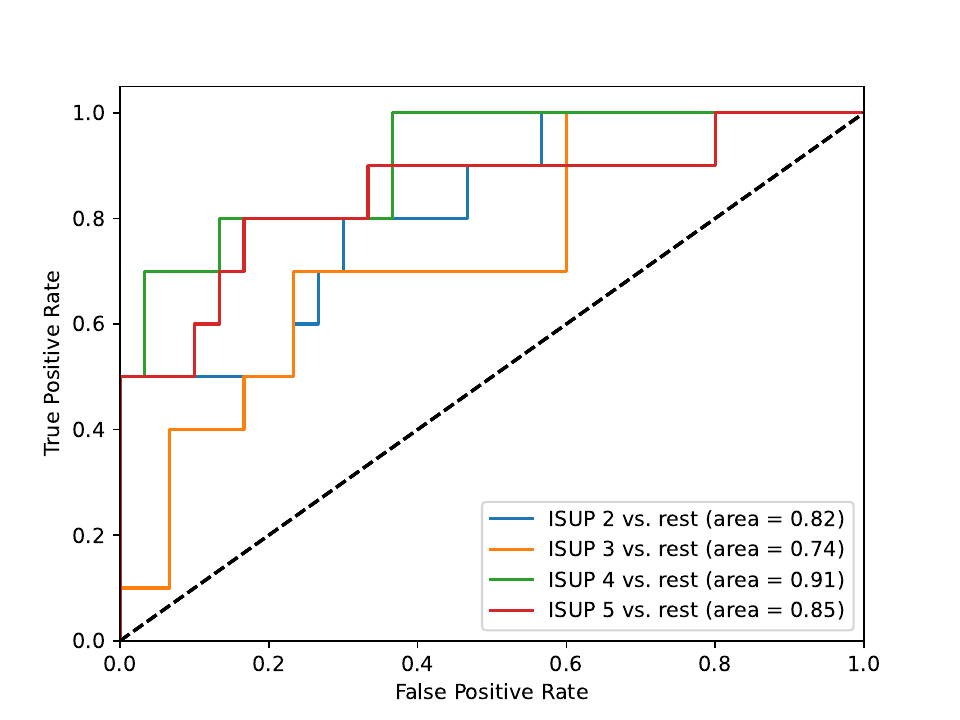}
\label{fig:picai_resnet50}}
\subfloat[VGG-16, PI-CAI dataset]{\includegraphics[width=0.25\textwidth]{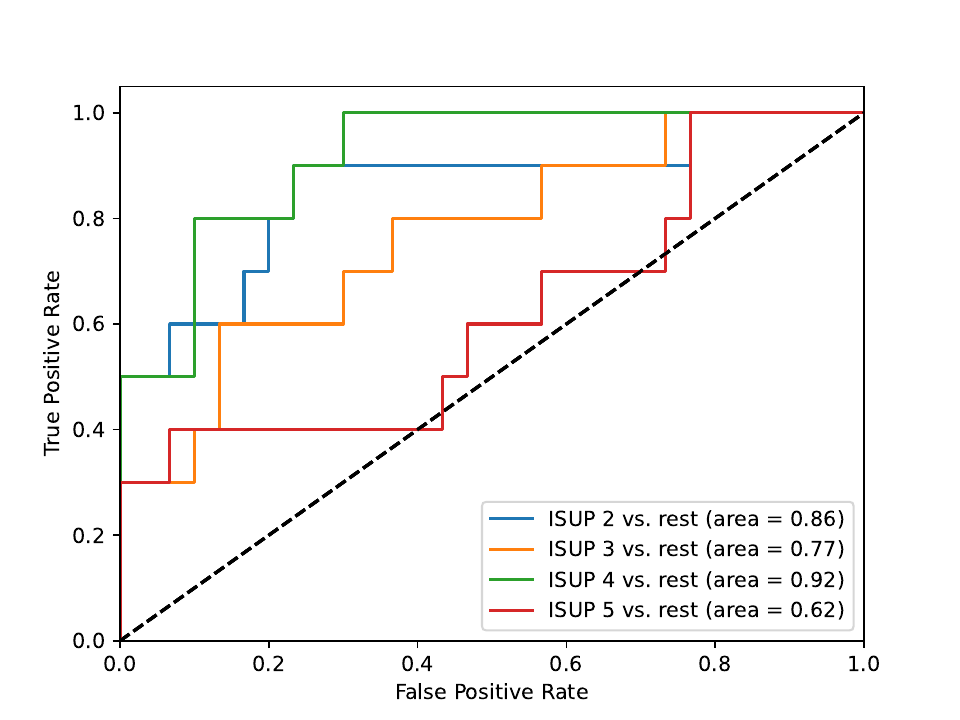}
\label{fig:picai_vgg16}}
\subfloat[DenseNet-121, PI-CAI dataset]{\includegraphics[width=0.25\textwidth]{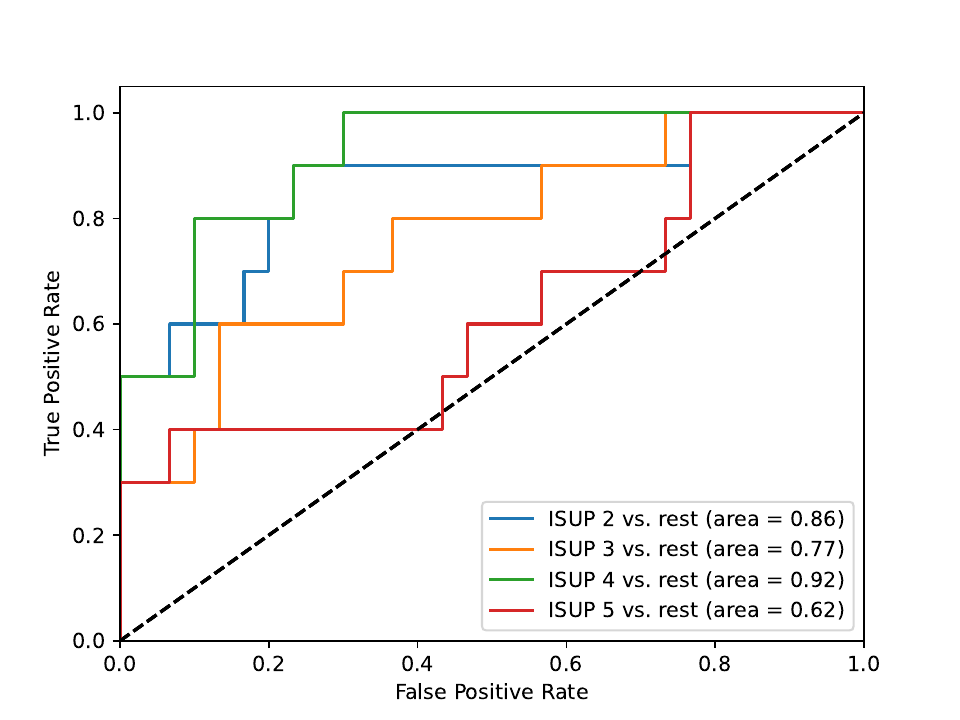}
\label{fig:picai_densenet121}}
\hfil
\subfloat[ResNet-18, BreakHis dataset]{\includegraphics[width=0.25\textwidth]{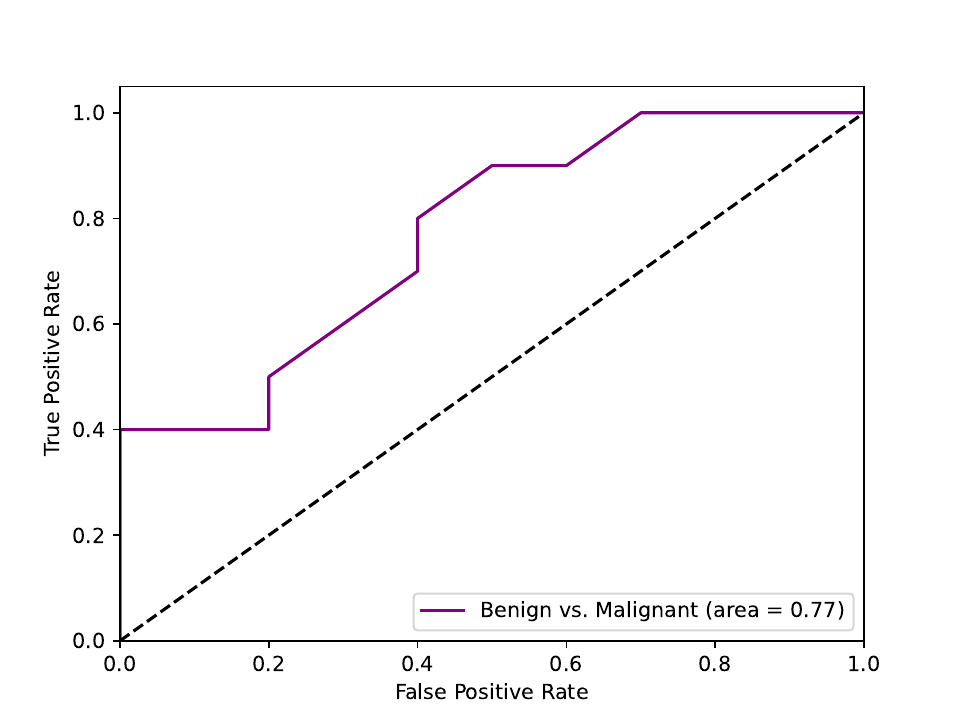}
\label{fig:breakhis_resnet18}}
%\hfil
\subfloat[ResNet-50, BreakHis dataset]{\includegraphics[width=0.25\textwidth]{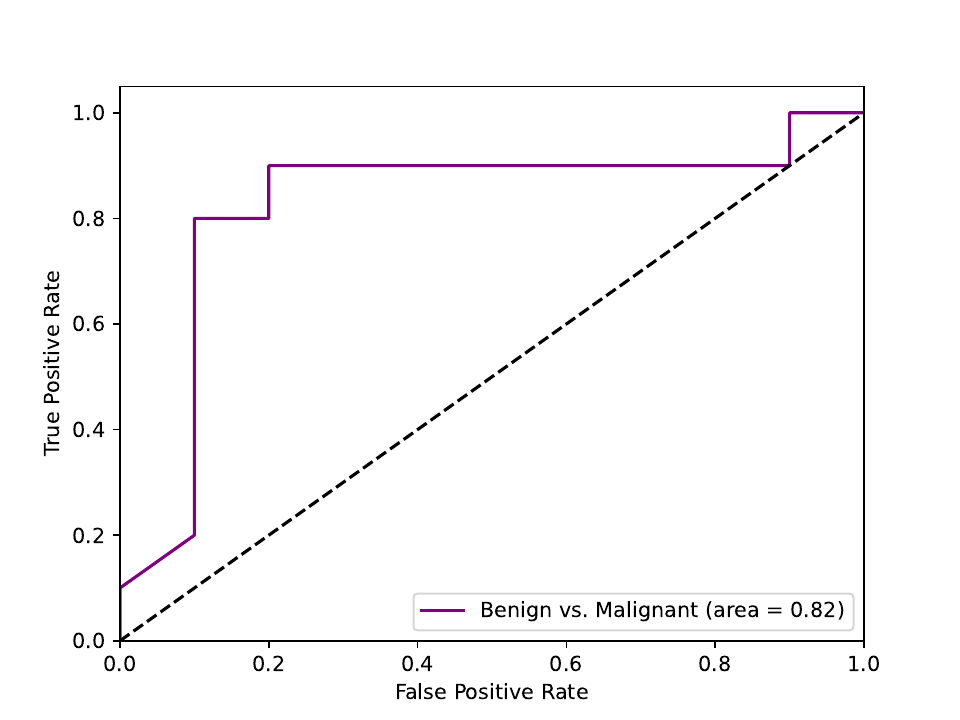}
\label{fig:breakhis_resnet50}}
\subfloat[VGG-16, BreakHis dataset]{\includegraphics[width=0.25\textwidth]{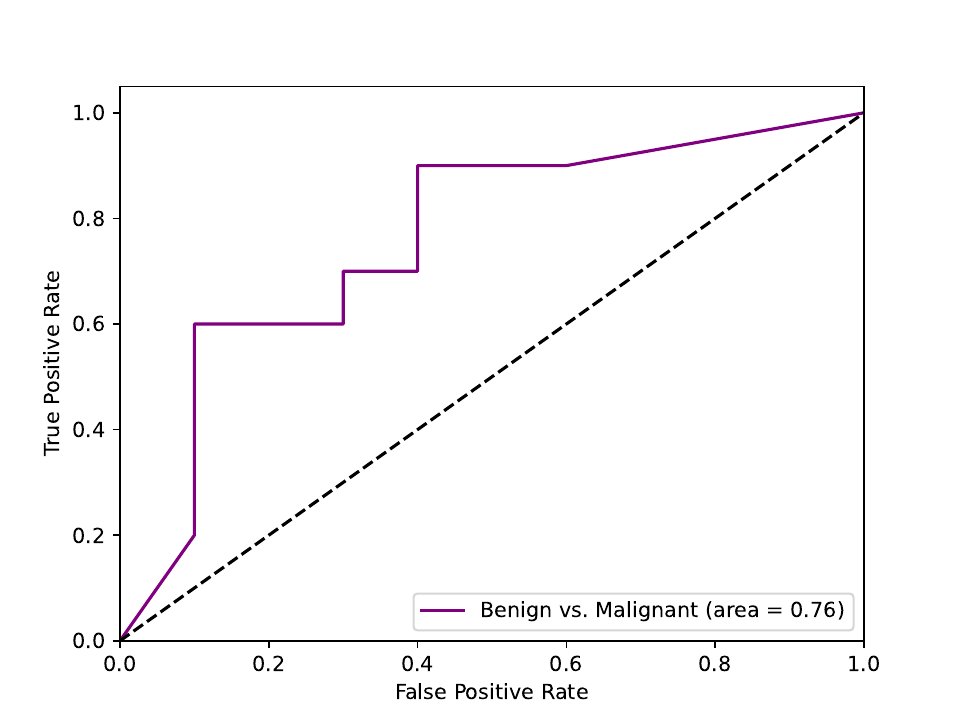}
\label{fig:breakhis_vgg16}}
\subfloat[DenseNet-121, BreakHis dataset]{\includegraphics[width=0.25\textwidth]{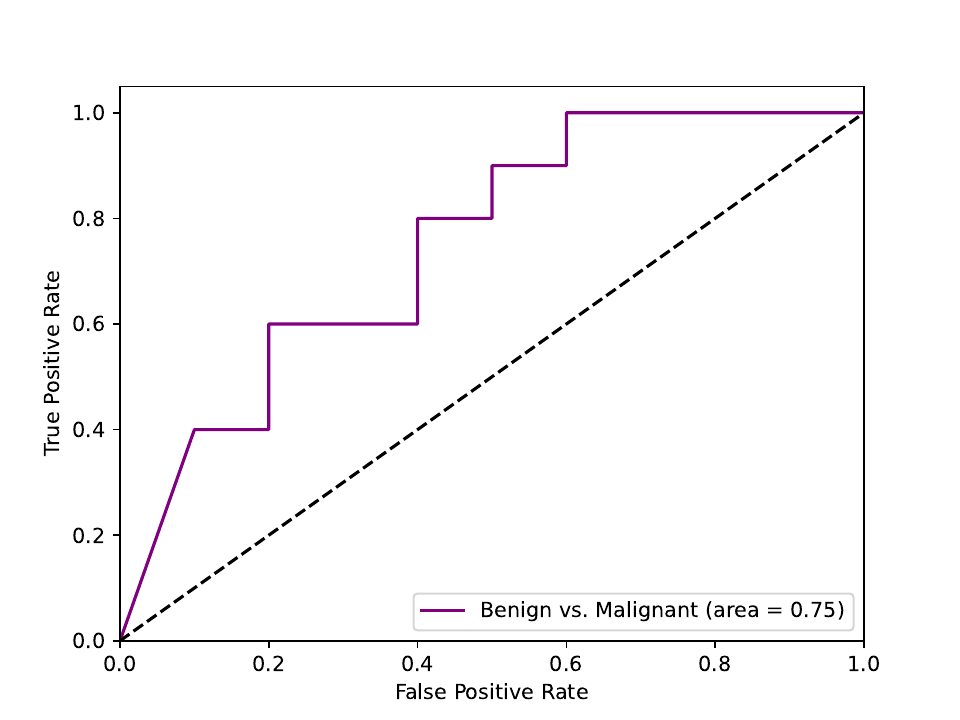}
\label{fig:breakhis_densenet121}}
\caption{ROC curves for one random episode of meta-testing of each backbone pre-trained with SimCLR+IP-IRM and meta-fine-tuned in a 5-shot setting. Meta-training and meta-testing classes have the same source dataset (PI-CAI or BreakHis).}
\label{fig:roc_curves}
\end{figure*}

Regarding the fine-tuning phase on the PI-CAI dataset, it is clear that, for all backbones, the fully-supervised approach establishes a lower bound, suggesting that the traditional training methodology alone is inadequate for this task. Indeed, in all the experiments, the proposed meta-fine-tuning scheme exceeds the fully-supervised performance even in a 1-shot setting. In particular, the proposed meta-learning scheme delivers the best-performing results in both 1-shot and 5-shot scenarios, achieving a peak classification performance of 0.780, 0.821, 0.803 and 0.771 multiclass AUROC for ResNet-18, ResNet-50, VGG-16 and DenseNet-121 models, respectively. For each backbone, we present the ROC curves for a randomly selected meta-testing episode of the best-performing combination of pre-training and meta-fine-tuning schemes in Fig.\ref{fig:picai_resnet18}, Fig.\ref{fig:picai_resnet50}, Fig.\ref{fig:picai_vgg16}, and Fig.\ref{fig:picai_densenet121}.

In contrast, for the BreakHis dataset, the fully-supervised model demonstrates strong generalization capabilities. This robust performance may be attributed to the lower complexity of the task compared to the classification of prostate cancer aggressiveness, stemming from several reasons. Firstly, the breast malignity classification task involves distinguishing only two classes in contrast to the four classes of the prostate case. Additionally, while microscopic images of breast cancer typically exhibit the presence of the disease throughout the image with cell nuclei indicating the presence of the tumor \cite{Young13} highlighted compared to background structures \cite{He12}, prostate MRI images depict lesions typically occupying only a few pixels within the image \cite{Bhattacharya22}. Furthermore, the lesion signal in prostate T2-weighted images may be isointense w.r.t. to the surrounding tissues \cite{Lovegrove18}, making its detection and categorization more challenging for the model. Lastly, in our PI-CAI experiments, we deliberately introduced an additional layer of complexity by explicitly considering different vendors between training and evaluation data. This aspect provides an additional level of complexity not present in the BreakHis dataset. Consequently, for the BreakHis dataset, the restricted data availability during meta-learning training tends to negatively impact the overall performance despite the enhanced generalization capabilities of the meta-learning framework.

Among the meta-learning experiments, the proposed meta-learning strategy performs best in both 1-shot and 5-shot settings, achieving peak performance with an AUROC of 0.772, 0.802, 0.791 and 0.797 for ResNet-18, ResNet-50, VGG-16 and DenseNet-121, respectively. We illustrate the ROC curves for a randomly chosen meta-testing episode of the best-performing pre-training and meta-finetuning combination for each backbone in Fig. \ref{fig:breakhis_resnet18}, Fig. \ref{fig:breakhis_resnet50}, Fig. \ref{fig:breakhis_vgg16} and Fig. \ref{fig:breakhis_densenet121}. Concerning the two ablation meta-learning experiments, for ResNet-18, VGG-16 and DenseNet-121, the performance trend is generally coherent, i.e., the performance degrades when considering a different meta-training source and especially when using a coarser class set for meta-training. The only exception is provided by the ResNet-50 backbone where, in the 5-shot setting, meta-training on coarser classes yields a higher AUROC (0.754) compared to meta-training on finer ones (0.739 AUROC).

Overall, our experiments demonstrate the effectiveness of our proposed methods,  which boost the FSL capabilities by embedding a disentangled SSL pre-training step and a novel meta-training scheme, outperforming most of the ablation experiments in both clinical challenges.
Specifically, the disentangled SSL pre-training step proves to be particularly effective as the training data increases, whereas, in a 1-shot setting, using vanilla SimCLR may sometimes provide better performance. Additionally, when the classification task is particularly challenging, as in the PI-CAI dataset, the proposed strategy significantly improves the classification capabilities of the model compared to a classical fully-supervised approach on the entire dataset. Furthermore, our results on the PI-CAI dataset demonstrate the strong generalization capabilities of the proposed approach even when training and evaluation data come from different vendors.
Either way, when the task is easier, as for the BreakHis dataset, and the fully-supervised baseline is stronger, our approach allows for achieving competitive classification capabilities in a few-data regime.
Finally, our findings demonstrate that even when meta-training and meta-testing data originate from different sources (unrelated meta-training and meta-testing classes), generating meta-training episodes from finer class sets contributes to improved generalization capabilities.

\section{Conclusions}
\label{sec:conclusion}
\noindent In this study, we proposed a novel method for boosting the capabilities of FSL-trained models by leveraging disentangled SSL and an enhanced meta-learning framework, and we demonstrated its effectiveness through extensive experiments and ablation studies on relevant medical imaging tasks. Furthermore, the diversity of clinical scenarios experimented demonstrated the versatility and potential broad application of our approach. Future work may delve into enriching the meta-learning algorithm's intrinsic capabilities by generating more informative prototypes leveraging support embeddings hallucination.

\vspace{5mm}

\noindent \textbf{CRediT authorship contribution statement} \\
%\vspace{2mm}
\normalsize \noindent \textbf{Eva Pachetti}: Conceptualization, Methodology, Software, Validation, Formal analysis, Investigation, Data Curation, Writing - Original Draft, Writing - Review \& Editing, Visualization. \textbf{Sotirios A. Tsaftaris}: Conceptualization,
Methodology, Writing - Review \& Editing, Supervision. \textbf{Sara Colantonio}: Conceptualization, Resources, Writing - Review \& Editing, Supervision, Project administration, Funding acquisition.

\vspace{5mm}

\noindent \textbf{Acknowledgments}\\
This study was funded by the European Union's Horizon 2020 Research and Innovation Program under Grant Agreement No 952159 (ProCAncer-I) and by the Regional Project PAR FAS Tuscany-NAVIGATOR. The funders had no role in the study design, data collection, analysis, interpretation, or manuscript writing.

\vspace{5mm}

\noindent\textbf{Declaration of Competing Interest} \\
The authors declare that they have no known competing financial
interests or personal relationships that could have appeared to influence
the work reported in this paper.

\bibliographystyle{unsrtnat}

\bibliography{bibliography}

\end{document}